\documentclass[conference]{IEEEtran}
\IEEEoverridecommandlockouts
\usepackage{cite}
\usepackage{amsmath,amssymb,amsfonts}
\usepackage{algorithmic}
\usepackage{graphicx}
\usepackage{textcomp}
\usepackage{hyperref}
\usepackage{xcolor}
\def\BibTeX{{\rm B\kern-.05em{\sc i\kern-.025em b}\kern-.08em
    T\kern-.1667em\lower.7ex\hbox{E}\kern-.125emX}}
\begin{document}

\title{A Comparative Study on Language Models for Task-Oriented Dialogue Systems\\
}

\author{
\IEEEauthorblockN{Vinsen Marselino Andreas}
\IEEEauthorblockA{
\textit{SEEI}\\
\textit{Institut Teknologi Bandung} \\
\textit{U-CoE AI-VLB} \\
Bandung, Indonesia \\
vinsenmarselino@gmail.com}
\and
\IEEEauthorblockN{Genta Indra Winata}
\IEEEauthorblockA{
\textit{ECE}\\
\textit{HKUST}\\
Clear Water Bay, Hong Kong \\
giwinata@connect.ust.hk}
\and
\IEEEauthorblockN{Ayu Purwarianti}
\IEEEauthorblockA{
\textit{SEEI}\\
\textit{Institut Teknologi Bandung} \\
\textit{U-CoE AI-VLB} \\
Bandung, Indonesia \\
ayu@informatika.org}
}

\IEEEoverridecommandlockouts
\IEEEpubid{\makebox[\columnwidth]{978-1-6654-1743-3/21/\$31.00 \copyright~2021 IEEE \hfill} \hspace{\columnsep}\makebox[\columnwidth]{ }}
\IEEEpubidadjcol

\maketitle

\begin{abstract}
    The recent development of language models has shown promising results by achieving state-of-the-art performance on various natural language tasks by fine-tuning pre-trained models. In task-oriented dialogue (ToD) systems, language models can be used for end-to-end training without relying on dialogue state tracking to track the dialogue history but allowing the language models to generate responses according to the context given as input. This paper conducts a comparative study to show the effectiveness and strength of using recent pre-trained models for fine-tuning, such as BART and T5, on end-to-end ToD systems. The experimental results show substantial performance improvements after language model fine-tuning. The models produce more fluent responses after adding knowledge to the context that guides the model to avoid hallucination and generate accurate entities in the generated responses. Furthermore, we found that BART and T5 outperform GPT-based models in BLEU and F1 scores and achieve state-of-the-art performance in a ToD system.
\end{abstract}



\section{Introduction}
Dialogue systems are developed to support human-to-human interactions in the natural language~\cite{Jurafsky2009}, and they are widely used in many applications, such as flight booking and hotel reservations. The task-oriented dialogue (ToD) systems commonly rely on modularized systems that use natural language understanding (NLU) to get input's meaning, separately with dialogue state tracking (DST) to track dialogue state and natural language generation (NLG) to generate suitable output. The benefit of applying this is the efficiency in training and inference during deployment. Recently,~\cite{madotto2020learning} show the possibility to utilize end-to-end models to replace the modularized systems, and they perform with decent performance.
To implement end-to-end ToD dialogue systems, there are two main ideas: (1) put knowledge base (KB) as input directly into the model \cite{madotto2018mem2seq}.
(2) develop a retrieval module to retrieve suitable knowledge from KB according to the input \cite{qin2019entity}. On the other line of work, \cite{madotto2020learning, wu2018end} utilized KB by augmenting samples using delexicalized templates. By applying this method, the trained model could learn the KB directly from the training dataset. By adding more datasets, the models can learn to utilize the knowledge in the context as input. \footnote{The code and dataset are available at~\url{https://github.com/sen33/end-to-end-dialogue-system}} The previous work has been focused on GPT-2 models as the pre-trained language models. However, there is no study yet on other language models, such as BART \cite{lewis2020bart} and T5 \cite{raffel2020exploring}. Both models are built with encoder-decoder architecture differently to GPT-2 that uses a decoder-only model. While, encoder-decoder models are utilized to develop end-to-end dialogue systems \cite{serban2016building, wen2017network}. The model accepts the input with the dialogue history and query and generates responses based on the context.

In this paper, we propose a comparative study to investigate the strength of language models for ToD systems. We also incorporate knowledge to the language models by two different methods: (1) applying Knowledge Embedded (KE) Dialogue~\cite{madotto2020learning} to leverage KB entities in delexicalized dialogue templates, and (2) adding KB in the input as context.
Our experiment shows that some language models perform better than others for end-to-end ToD systems. We found that the models with pre-trained models produce more fluent responses after incorporating knowledge embedded. Furthermore, we found that BART and T5 outperform GPT-2-based models in both BLEU and F1 scores and achieve state-of-the-art performance in the CamRest dataset~\cite{wen2017network}.

\IEEEpubidadjcol

\section{Methodology}
In this section, we would describe the task of an end-to-end task-oriented dialogue system and how we prepare the dataset.

\subsection{Notation and Task}
We define a dialogue dataset $\mathcal{D} = \{D_1, D_2, \cdots, D_n\}$ where each dialogue has user and system utterances or alternating dialogue turns $D_i = \{U_1, S_1, U_2, S_2, \cdots, U_t, S_t\}$. For each dialogue sample, we define a query $Q$ and dialogue history $H$. In the end-to-end dialogue system, we define our generative model as $\theta$. The model $\theta$ takes $x = [H, Q]$ as a concatenation of the dialogue history $H$ and query $Q$ as input, and generates an output response $y$. The dialogue history $H$ is taken from the previous turns of the query $Q$. We fine-tune $\theta$ using the dialogue samples by the conditional generation objective. It trains the model by conditioning to the context. We define the loss as the following:
\begin{align}
    \mathcal{L}(\mathcal{D}) = - \sum_{i}^n \log p_{\theta}(s_i|s_{i<1}, x_i),
\end{align}
where $p_{\theta}(s_i|s_{<i}, x_i)$ is the conditional probability of generating token $s_i$ given the previous tokens $s_{<i}$ and the context $x_i$. On the inference time, greedy search are used to generate the response. 

\subsection{Generative Language Models}
In this section, we describe the models that are used in this work as the following:
\subsubsection{Sequence-to-sequence}
As our baseline model, we train vanilla encoder-decoder models using transformer with multi-head attention~\cite{vaswani2017attention} using OpenNMT toolkit~\cite{klein-etal-2017-opennmt}. This toolkit has been widely used for training sequence-to-sequence on NLP tasks~\cite{muis2020sequence}.

\subsubsection{Fine-tuning using Pre-trained Models}
\paragraph{Bidirectional and Auto-regressive Transformers (BART)}
BART~\cite{lewis2020bart} is a language model that is trained using the masked language modeling from BERT~\cite{devlin2019bert}, and denoising objective to recover the perturbed input.

\paragraph{Text-to-Text Transfer Transformer (T5)}
T5 is an encoder-decoder based language models~\cite{raffel2020exploring}. This model is trained using BERT~\cite{devlin2019bert} training objective by applying the mask to the input tokens.

\subsubsection{Embed Knowledge}
To evaluate the effectiveness of adding KB information in the end-to-end ToD systems, we incorporate the knowledge in two ways: (1) the KE Dialogue data augmentation method, and (2) Adding the KB in the context as input, shown in Table~\ref{HistoryTargetExample}.

\subsection{Dataset Preparation}
\begin{figure}[!ht]
\centerline{\includegraphics{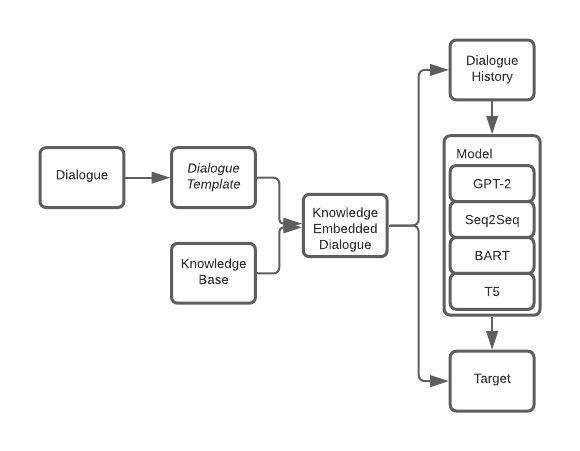}}
\caption{The training and evaluation step of KE Dialogue~\cite{madotto2020learning}.}
\label{Flow}
\end{figure}

Dataset is prepared by following the KE Dialogue \cite{madotto2020learning}.
Fig~\ref{Flow} shows the overall flow of the system. The dialogue template is extracted from each dialogue by delexicalization (\texttt{KE-DELEX}) using the entities from the dialogue ontology. Then, the templates are embedded with entities from knowledge bases to form knowledge embedded dialogue by relexicalization (\texttt{KE-RELEX}). For every question and answer, one dialogue history and target will be generated.
To create the representation of dialogue history, each sentence is concatenated with a special token separator.
\texttt{\(\langle\)USR\(\rangle\)} token is concatenated before user's sentence, and \texttt{\(\langle\)SYS\(\rangle\)} token is concatenated before system's sentence.
These pairs of dialogue history and target will be the input and target of the trained models. To add KB directly into input, special token \texttt{\(\langle\)DTA\(\rangle\)} is concatenated before every entity available from the intermediate API.
The example of dialogue history with KB as input is shown in Table~\ref{HistoryTargetExample}.

\begin{table}[htbp]
\caption{Example of dialogue history with KB.}
\begin{center}
\begin{tabular}{|p{1.2cm}|p{7cm}|}
\hline
\textbf{Input} & \(\langle\)USR\(\rangle\) i would like a moderately priced restaurant in the north part of town . \(\langle\)SYS\(\rangle\) golden\_wok is a moderately priced restaurant in the north side of town . \(\langle\)USR\(\rangle\) what type of food does golden\_wok serve ? \(\langle\)DTA\(\rangle\) the\_nirala 7\_milton\_road\_chesterton north indian 52.215157,0.125015 01223\_360966 moderate cb41uy \(\langle\)DTA\(\rangle\) golden\_wok 191\_histon\_road\_chesterton north chinese 52.220757,0.111564 01223\_350688 moderate cb43hl \\ \hline
\textbf{Target} & the golden\_wok serves chinese food . would you like more information ? \\ \hline
\end{tabular}
\label{HistoryTargetExample}
\end{center}
\end{table}

\section{Experiment}

\subsection{Dataset}
We use CamRest dataset\cite{wen2017network}, a human-to-human dialogues dataset for restaurant recommendation in Cambridge.
We use a dataset that is already preprocessed and use the code provided by \cite{madotto2020learning} to extract 161 template dialogues and generate Knowledge Embedded Dialogues. 676 dialogues are provided by the CamRest dataset. It is split into 406, 135, and 135 as training data, validation data, and test data, respectively.
The templates are generated from training data and augment 9,728 new dialogues to the training data.

\subsection{Model Configuration}
For each pre-trained model (BART and T5), there are four hyper-parameters configurations. Each configuration is a combination of batch sizes [8, 16] and learning rate [1e-5, 1e-4].
For BART, we use BART$_{\text{BASE}}$ that has 12 layers, attention heads of 16, 3,072 feed-forward, and 768-dimension embeddings with 139M parameters is used. For T5, we use T5$_{\text{BASE}}$ that has 24 layers, 12 attention heads, 3,072 feed-forward, and 768-dimension embeddings with 220M parameters is used.
Every configuration use 30 epoch with the early stopping method.
Early stopping evaluates the BLEU score from the validation dataset every epoch and picks the best one.
For sequence-to-sequence with OpenNMT, there are two hyper-parameter configurations with two model sizes (small and large).
All Seq2seq configurations use transformer encoder and decoder.
Table~\ref{SmallLargeModel} shows the number of parameters of these Seq2seq models.
All experiments were conducted on Tesla V100 GPU machines. Adam optimizer is used and learning rate is updated using a linear scheduler.

\begin{table}[!ht]
\caption{Parameter for small and large Seq2Seq model}
\begin{center}
\begin{tabular}{|c|c|c|}
\hline
\bfseries{Parameter} & \bfseries{Small model} & \bfseries{Large model} \\ \hline
Step & 100k & 50k \\ \hline
Batch size & 8 & 16 \\ \hline
Learning rate & 6.25e-5 & 6.25e-5\\ \hline
Layer & 12 (6 enc, 6 dec) & 12 (6 enc, 6 dec) \\ \hline
Attention head & 8 & 8 \\ \hline
Feedforward & 1024 & 3072 \\ \hline
Embedding & 512 & 768 \\ \hline
\end{tabular}
\label{SmallLargeModel}
\end{center}
\end{table}

\section{Results and Analysis}
In this section, we report the results, analyze our findings and ablation study, and conduct a human evaluation to measure the quality of our model's responses.

\subsection{Results}

\begin{table}[htbp]
\caption{Results for pre-trained model with different hyper-parameters.}
\begin{center}
\begin{tabular}{|c|c|c|c|c|}
\hline
\bfseries{Model} & \bfseries{Batch size} & \bfseries{Learning rate} & \bfseries{BLEU} & \bfseries{F1} \\ \hline
BART$_{\text{BASE}}$ & 8 & 1e-5 & \bfseries{19.740} & 46.036 \\ \hline
\bfseries{BART$_{\text{BASE}}$} & 16 & 1e-5 & 19.050 & 55.922 \\ \hline
BART$_{\text{BASE}}$ & 8 & 1e-4 & 18.240 & \bfseries{56.202} \\ \hline
BART$_{\text{BASE}}$ & 16 & 1e-4 & 17.930 & 51.423 \\ \hline
\hline
T5$_{\text{BASE}}$ & 8 & 1e-5 & 18.140 & 53.301 \\ \hline
T5$_{\text{BASE}}$ & 16 & 1e-5 & 16.490 & 49.927 \\ \hline
T5$_{\text{BASE}}$ & 8 & 1e-4 & 18.330 & 56.187 \\ \hline
\bfseries{T5$_{\text{BASE}}$} & 16 & 1e-4 & \bfseries{18.730} & \bfseries{56.311} \\ \hline
\end{tabular}
\label{ResultAllModel}
\end{center}
\end{table}

The result of the pre-trained model is shown in Table~\ref{ResultAllModel}.
For sequence-to-sequence, the smaller model achieves a better BLEU and F1 score. For BART model, the best model was achieved by a model that use batch size of 16 with learning rate of 1e-5. 
However, the difference between the BLEU and F1 with the best model in each metrics is marginal. For T5 model, the best model is the model with batch size of 16 with learning rate of 1e-4.
This model achieves the best score in both BLEU and T5 compared to other T5 models.

\begin{table}[htbp]
\caption{Result for best model configuration.}
\begin{center}
\begin{tabular}{|c|c|c|c|}
\hline
\bfseries{Model} & \bfseries{Parameter} & \bfseries{BLEU} & \bfseries{F1} \\ \hline
Seq2Seq$_{\text{SMALL}}$ & 33M & 17.870 & 49.304 \\ \hline
Seq2Seq$_{\text{LARGE}}$ & 101M & 16.220 & 45.438 \\ \hline
BART & 139M & \textbf{19.050} & 55.922 \\ \hline
T5 & 220M & 18.730 & \textbf{56.311} \\ \hline
\end{tabular}
\label{ResultBestModel}
\end{center}
\end{table}

We show the performance of our best models for BART and T5 in Table~\ref{ResultBestModel}.
Both BART and T5 best model use batch size of 16. While BART achieves better BLEU, T5 achieves a better F1 score.
This is caused by how the model was pre-trained. BART is pre-trained by denoising sequence, so the model achieves better BLEU, a metric that shows how fluent the predictions are. Using BART and T5 models for initialization outperform the vanilla sequence-to-sequence model. It implies that pre-trained models have learned the knowledge that are useful for building ToD systems.

\begin{table}[htbp]
\caption{Comparison of results with existing works.}
\begin{center}
\begin{tabular}{|c|c|c|}
\hline
\bfseries{Model} & \bfseries{BLEU} & \bfseries{F1} \\ \hline
\multicolumn{1}{|r|}{KB-Transformer \cite{haihong2019kb}} & 14.80 & 45.30 \\ \hline
\multicolumn{1}{|r|}{MLMN \cite{reddy2019multi}} & 13.61 & 54.85 \\ \hline
\multicolumn{1}{|r|}{BoSsNet \cite{raghu2019disentangling}} & 15.20 & 43.10 \\ \hline
\multicolumn{1}{|r|}{KB-Retriever \cite{qin2019entity}} & 18.64 & 55.76 \\ \hline
\hline
\multicolumn{1}{|r|}{GPT-2 \cite{madotto2020learning}} & 13.58 & 34.69 \\ \hline 
\multicolumn{1}{|r|}{GPT-2+KB \cite{madotto2020learning}} & 13.59 & 50.45 \\ \hline 
\multicolumn{1}{|r|}{GPT-2+KE \cite{madotto2020learning}} & 18.00 & 54.85 \\ \hline \hline
\multicolumn{1}{|r|}{Seq2Seq+KE} & 17.870 & 49.304 \\ \hline
\multicolumn{1}{|r|}{BART+KE} & \bfseries{19.050} & 55.922 \\ \hline
\multicolumn{1}{|r|}{T5+KE} & 18.730 & \bfseries{56.311} \\ \hline
\end{tabular}
\label{ResultComparison}
\end{center}
\end{table}

We show the comparison of our models with the best hyper-parameters setting in~Table~\ref{ResultComparison}. Seq2Seq models achieve worse performance compared to some baselines, especially \texttt{GPT-2+KE}.
Both BART and T5 achieve higher BLEU and F1 scores compared to all baselines.

\begin{table}[htbp]
\caption{Ablation study without KB, with KB, and with KE.}
\begin{center}
\begin{tabular}{|c|c|c|}
\hline
\bfseries{Model} & \bfseries{BLEU} & \bfseries{F1} \\ \hline
\multicolumn{1}{|r|}{BART} & 19.100 & 41.580 \\ \hline
\multicolumn{1}{|r|}{BART+KB} & \textbf{20.240} & \textbf{56.704} \\ \hline
\multicolumn{1}{|r|}{BART+KE} & 19.050 & 55.922 \\ \hline
\end{tabular}
\label{BARTKEKB}
\end{center}
\end{table}

\subsection{Ablation Study}
To compare the effectiveness of applying KE and KB to the language model, ablation study is conducted. Initially, the BART model is chosen as the base model in the experiment. Then, aside from using \texttt{BART+KE}, we also train a model without any augmentation (BART) and a model using KE (\texttt{BART+KB}).
For \texttt{BART+KB}, every entity from intermediate API is concatenated to dialogue history with a special token \texttt{\(\langle \text{DTA} \rangle\)}.
The result is shown in Table~\ref{BARTKEKB}.
\texttt{BART} achieves better BLEU than \texttt{BART+KE} by a slight margin but falls behind in the F1 score to \texttt{BART+KB} and \texttt{BART+KE}.
It means that adding KB directly into input or with KE reduces hallucination, a condition where the generated sequence has good structure and meaning but the wrong entity.

\subsection{Human Evaluation}
Human evaluation is done for \texttt{BART+KB} and \texttt{BART+KE} to further measure the humanness of our generation results.
A Likert scale \cite{allen2007likert} of 1, 3, and 5 are given to all test predictions by experts.
Table~\ref{HumanEvaluation} shows the result of this evaluation.
It shows that by using KE Dialogues as training data, the trained model is more robust and more human-like, as demonstrated in Table~\ref{HumanEvaluation}.

\begin{table}[htbp]
\caption{Human evaluation for BART+KB and BART+KE.}
\begin{center}
\begin{tabular}{|c|c|}
\hline
\bfseries{Model} & \bfseries{Likert Score} \\ \hline
\multicolumn{1}{|r|}{BART+KB} & 3.76 \\ \hline
\multicolumn{1}{|r|}{BART+KE} & \textbf{4.14} \\ \hline
\end{tabular}
\label{HumanEvaluation}
\end{center}
\end{table}

The example of input and output is shown in Table~\ref{OutputExample}. Each model could generate an answer that is understood by a human. Models tend to directly suggest a restaurant's name instead of asking for specific information.

\begin{table}[htbp]
\caption{The examples of the dialogue input and output on different models.}
\begin{center}
\begin{tabular}{|p{1.2cm}|p{7cm}|}
\hline
\bfseries{Input} & \bfseries{\(\langle\)USR\(\rangle\) i am looking for a restaurant that is in the expensive price range and in the south part of town} \\ \hline
Target & there are results matching your query . would you like mexican , italian , chinese , or indian ? \\ \hline
Seq2Seq & the\_good\_luck\_chinese\_food\_takeaway serves expensive food in the south part of town . \\ \hline
BART & peking\_restaurant serves \textbf{expensive food} in the \textbf{south part of town}. \\ \hline
T5 & taj\_tandoori serves \textbf{expensive food} in the \textbf{south part of town}. \\ \hline \hline
\bfseries{Input} & \bfseries{\(\langle\)USR\(\rangle\) i am looking for a restaurant that is in the expensive price range and in the south part of town . \(\langle\)SYS\(\rangle\) there are results matching your query . would you like mexican , italian , chinese , or indian ? \(\langle\)USR\(\rangle\) let 's go with italian food .} \\ \hline
Target & frankie\_and\_bennys is an expensive italian eatery in the south part of town . would you like any additional information about this restaurant ? \\ \hline
Seq2Seq & frankie\_and\_bennys is an \textbf{expensive restaurant} in the \textbf{south part of town} . \\ \hline
BART & frankie\_and\_bennys is an \textbf{italian restaurant} in the \textbf{south part of town}. \\ \hline
T5 & frankie\_and\_bennys serves \textbf{italian food} in the \textbf{south part of town}. is there anything else i can help you with? \\ \hline
\end{tabular}
\label{OutputExample}
\end{center}
\end{table}

\section{Related Work}
The first task-oriented dialogue system is \texttt{ELIZA} \cite{weizenbaum1966eliza}, a dialogue system that utilize parsers and rule-based engines. Then, \cite{young2013pomdp} explored developing dialogue systems by utilizing statistical-based methods using POMDP. Along with the development of machine learning, deep learning received a lot of attention from researchers to develop models on modularized dialogue systems, such as NLU~\cite{hakkani2016multi,chen2016end,goo2018slot,liu2020attention}, DST~\cite{wu2019transferable,lin2020mintl}, and NLG start to utilize deep learning approaches.
The specificity of modularized dialogue systems leads to an idea where the DST module is bypassed, which is end-to-end dialogue systems.
Handling new domains could be achieved by end-to-end dialogue systems with retraining the model, unlike modularized dialogue systems that need to change the DST.
To handle KB in the end-to-end dialogue systems, there are two main ideas, using KB directly as input \cite{madotto2018mem2seq} or using intermediate API to retrieve correct KB \cite{qin2019entity}.
\cite{madotto2020learning} propose another idea where KBs are embedded into dialogue templates to form KE Dialogue and achieve promising results.

\section{Conclusion}
This paper shows the effectiveness of applying pre-trained language models for fine-tuning end-to-end task-oriented dialogue systems and incorporating knowledge bases as context. Using pre-trained language models is essential for initialization to improve the generation results in terms of fluency. Moreover, adding KB to the context improves the correctness by reducing the hallucination. We found that BART and T5 models achieve state-of-the-art performance with higher BLEU and F1 scores compared to GPT-2 models with very similar sizes.

\section*{Acknowledgment}
This research is partially funded by 
Center for Artificial Intelligence of Institut Teknologi Bandung.

\bibliographystyle{IEEETran}
\bibliography{references.bib}
\vspace{12pt}

\end{document}